\documentclass[conference]{IEEEtran}
\ifCLASSINFOpdf
\else
\fi

\usepackage{color}
\usepackage{algorithm,algorithmicx}
\usepackage{algpseudocode,empheq,hhline}
\usepackage{amsmath,amsthm,amssymb,url,graphicx,rotating,ifthen,epsfig,array,caption,color}
\usepackage[bookmarks=false]{hyperref}
\usepackage{todonotes}
\usepackage{graphicx}
\usepackage{caption}
\usepackage{subcaption}
\usepackage{balance}
\usepackage{multirow}

\makeatletter
\newcommand{\putindeepbox}[2][0.7\baselineskip]{{%
    \setbox0=\hbox{#2}%
    \setbox0=\vbox{\noindent\hsize=\wd0\unhbox0}
    \@tempdima=\dp0
    \advance\@tempdima by \ht0
    \advance\@tempdima by -#1\relax
    \dp0=\@tempdima
    \ht0=#1\relax
    \box0
}}
\makeatother

\newcounter{eqn}

\DeclareMathOperator*{\argmax}{argmax}
\def\N{{\mathbb N}}
\def\E{{\mathbb E}}
\def\P{{\mathbb P}}
\def\G{{\cal N}}
\def\U{{\cal U}}
\def\R{{\mathbb R}}
 
\def\s{\mathbf{s}}
\begin{document}
%
\title{Bandit-Based Random Mutation Hill-Climbing}

\author{
\IEEEauthorblockN{Jialin Liu}
\IEEEauthorblockA{University of Essex\\
Colchester CO4 3SQ\\
United Kingdom\\
\url{jialin.liu@essex.ac.uk}}
\and
\IEEEauthorblockN{Diego P\'erez-Li\'ebana}
\IEEEauthorblockA{University of Essex\\
Colchester CO4 3SQ\\
United Kingdom\\
\url{dperez@essex.ac.uk}}
\and
\IEEEauthorblockN{Simon M. Lucas}
\IEEEauthorblockA{University of Essex\\
Colchester CO4 3SQ\\
United Kingdom\\
\url{sml@essex.ac.uk}}
}


%


\maketitle

\begin{abstract}
The Random Mutation Hill-Climbing algorithm is a direct search technique mostly used in discrete domains. It repeats the process of randomly selecting a neighbour of a best-so-far solution and accepts the neighbour if it is better than or equal to it. In this work, we propose to use a novel method to select the neighbour solution using a set of independent multi-armed bandit-style selection units which results in a bandit-based Random Mutation Hill-Climbing algorithm. The new algorithm significantly outperforms Random Mutation Hill-Climbing in both OneMax (in noise-free and noisy cases) and Royal Road problems (in the noise-free case). The algorithm
shows particular promise for discrete optimisation problems where each fitness evaluation is expensive.
\end{abstract}


%


\section{Introduction}

Evolutionary Algorithms (EA) have achieved widespread use since their developments in the 1950s and 1960s~\cite{box1957evolutionary,friedman1959digital,bledsoe1961use,bremermann1962optimization,rechenberg1965cybernetic,fogel1966artificial,reed1967simulation}.

Their essence is relatively simple: to generate an initial set of candidate solutions at random, and then to iteratively improve the candidate set via a process of variation, evaluation and selection. They have been the subject of much analysis, development and a diverse range of applications.  They have also spawned related approaches (i.e. methods which can be characterised  by the outline description above) such as particle swarm optimisation, and have been extended for application to multi-objective optimisation.

This paper introduces a significant variation: the Bandit-Based Evolutionary Algorithm.
Bandit algorithms~\cite{gittins1979bandit,berry1985bandit} have become popular for optimising either simple regret (the best final decision after a number of exploratory trials) or cumulative regret (best sum of rewards over a number of trials) in A/B testing.

A popular bandit algorithm is the Upper Confidence Bound (UCB) algorithm~\cite{lairobbins,auer2002finite} which balances the trade-off between exploration and exploitation.
The UCB-style algorithms have achieved widespread use within Monte Carlo Tree Search (MCTS)~\cite{browne2012survey}, called UCT when applied to trees, the ``T'' being for Trees.

A wide literature exists on bandits~\cite{lairobbins,auer2002finite,Auer03,audibert2006use,bubeck2009pure,garivier2011kl,bubeck2012regret} and many tools have been proposed for distributing the computational power over the stochastic arms to be tested.
There are also some adaptations to other contexts: time varying as in \cite{kocsis2006discounted}; adversarial~\cite{grigoriadis,auer95gambling}); or involving the non-stationary nature of bandit problems in optimization portfolios.
St-Pierre and Liu~\cite{stpierre:hal-00979456} applied the Differential Evolution algorithm~\cite{storn1997differential} to some non-stationary bandit problem, which outperformed the classical bandit algorithm on the selection over a portfolio of solvers.

Browne et al.~\cite{browne2012survey} noted the great potential for hybridising MCTS with other approaches to optimisation and learning, and in this paper we provide a hybridisation of an evolutionary algorithm with a bandit algorithm.

There are examples of using evolution to tune MCTS parameters~\cite{benbassat2013evolving,alhejali2013using,lucas2014fast}. Albeit robust, this application of EA is not widespread, due to the computational cost involved in performing fitness evaluations. It should be noted that Lucas et al.~\cite{lucas2014fast} made fitness evaluations after each rollout, so they could be rapidly optimised, albeit noisily.

The algorithm reported in this paper is a very different hybrid: it uses bandits to represent the state of the evolving system. This has some similarities with Estimation of Distribution Algorithms (EDAs) \cite{rolet2010bandit} but the details are significantly different.

To our knowledge, this is one of the very few times that this type of hybridisation has been attempted; the only other paper we are aware of in the same vein is Zhang et al.~\cite{li2014adaptive}. Zhang et al. used a bandit algorithm as a form of Adaptive Operator Selection: the variation operators used within the evolutionary algorithm were selected using a bandit-based approach, showing promising results.

In this paper we develop a bandit-based version of the Random Mutation Hill-Climbing (RMHC) algorithm, and compare the two methods, i.e., the original and the bandit-based algorithms.  

To put this in some context, it should be noted that while the RMHC algorithm is very simple, it is often surprisingly competitive with more complex algorithms, especially when deployed with random restarts.

For instance, Lucas and Reynolds evolved Deterministic Finite Automata (DFA)~\cite{lucas2003learning,lucas2005learning}, using
a multi-start RMHC algorithm with very competitive results,
outperforming more complex evolutionary algorithms, and for some
classes of problems also outperforming the state of the art Evidence-Driven State Merging (EDSM) algorithms. 

Although Goldberg~\cite{goldberg1989genetic} used bandit models, they were used to help understand the operation of a Simple Genetic Algorithm.
Our approach is different: we use them as the very basis of the algorithm. The bandit model provides a natural way to balance exploitation (sticking with what appears to be good) versus exploration (trying things which have not been sampled much).

In this paper we model the genome as an array of bandits. In the case where the genome is a binary string of length $n$, we model the state of the evolutionary system as an array of $2$-armed bandits. If each element of the string can take on $m$ possible values, then each element is represented by an $m$-armed bandit.

The main contribution in this work is this new bandit-based RMHC algorithm, together with results on some standard benchmark problems. The new algorithm significantly outperforms the standard RMHC in both OneMax (in noise-free and noisy cases) and Royal Road problems in the noise-free case. Tests on noisy Royal Road problems will be studied in future work.

\section{Test Problems}
In this work, we consider two benchmark optimisation problems in a binary search space.

\subsection{OneMax Problem}
The OneMax problem~\cite{schaffer1991crossover} is a simple linear problem aiming at maximising the number of $1$ of a binary string, i.e., for a given $n$-bit string $\mathbf{s}$
\begin{equation}
f(\s)=\sum_{i=1}^{n} s_i,
\end{equation}
where $s_i$, denoting the $i^{th}$ bit in the string $\mathbf{s}$, is either 1 or 0.

The complexity of OneMax problem is $O(n\log(n))$ for a $n$-bit string~\cite{droste2006upper}. Doerr et al. proved that the black-box complexity with memory restriction one is at most $2n$~\cite{doerr2012memory}. More lower and upper bounds of the complexity of OneMax in the different models are analysed~\cite{doerr2013black,Doerr2016} and then summarised in Table 1 of \cite{Doerr2016}. In their elitist model, only the best-so-far solution can be kept in the memory.
Our bandit-based RMHC stores the best-so-far solution in the noise-free environment and stores additionally its evaluation number in the noisy environment (detailed in Section \ref{sec:noisy}).

\subsection{Noisy OneMax Problem}
We modify the OneMax problem by introducing an additive noise with constant variance $1$:
\begin{equation}
f'(\s)=f(\s)+ \G(0,1),
\end{equation}
$\G$ denotes a Gaussian noise.
Thus, the noise standard deviation is of same order as the differences between fitness values. It is notable that our noise model is very different from the one in \cite{qian2015analyzing}, which used (1+1)-EA and a one-bit noise.

The influence of the noise strength on the runtime of (1+1)-EA for the OneMax problem corrupted by one-bit noise is firstly analysed by Droste~\cite{droste2004analysis}. In \cite{droste2004analysis} and \cite{qian2015analyzing}, the misranking occurs due to the change of exactly one uniformly chosen bit of $\s$ by noise with probability $p \in (0,1)$, $p$ is the noise strength. Thus, the noise acts before fitness evaluation and the evaluated individual (solution or search point) is possibly not the correct one. The individual is infected by noise in their model while in our model, the fitness function is infected by noise.

\subsection{Royal Road Function}
The Royal Road functions are firstly introduced by Mitchell et al.~\cite{mitchell1992royal}. The function fitness gains only if all the bits in one block are flipped to $1$.\footnote{Note that OneMax can be considered as a special case of the Royal Road function with a block size of $1$.}
The objective was to designing some hierarchical fitness landscapes and studying the performance of Genetic Algorithms (GA).
Surprisingly, a simple Random Mutation Hill-Climbing Algorithm outperforms GA on a simple Royal Road function, namely $R_1$ in \cite{mitchell1992royal,mitchell1993will}.
$R_1$ consists of a list of block-composite bit strings as shown in Fig. \ref{fig:royal}, in which `*' denotes a either 0 or 1.
The fitness $R_1(\mathbf{x})$ is recalled as follows:
\begin{equation}\label{eq:r1}
R_1 = \sum_{i=1}^{n} c_i \delta_i(\mathbf{x}),
\end{equation}
where $\delta_i=1$ if $\mathbf{x}\in \s_i$, otherwise, $\delta_i=0$. 
\begin{figure}[h]
\centering
\includegraphics[width=0.8\linewidth]{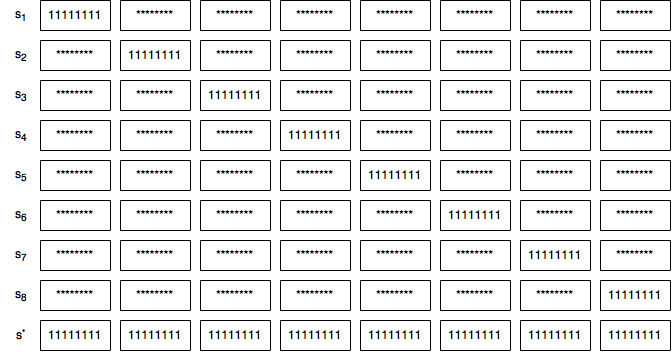}
\caption{\label{fig:royal}Royal Road function $R_1$.}
\end{figure}

It's notable that, due to the landscapes in the Royal Road function and 1-bit mutation per generation, introducing noise to the fitness is not trivial.
When introducing a noise with constant variance, with high probability, the mutated genome has an identical noise-free fitness value to the one of its ancestor. As a result, only the samples of introduced noise are compared.

\def\later{
\subsection{Noisy Royal Road Function}
We introduce the same addictive noise to the Royal Road function as to the OneMax problem:
\begin{equation}
R'_1(\s)=R_1(\s) + \G(0,1).
\end{equation}

It's notable that, due to the landscapes in the Royal Road function and 1-bit mutation per generation, introducing noise to the fitness is not trivial.
When introducing a noise with constant variance, with high probability, the mutated genome has an identical noise-free fitness value to the one of its ancestor. As a result, only the samples of introduced noise are compared.
The noisy Royal Road is a very difficult and meaningful problem.
}

\section{Bandit-based RMHC}

In contrast to the standard bandit terminology, where an arm is pulled to gain some reward, 
the purpose of our bandits is to select the element to mutate at each iteration of the algorithm.  

We create an $m$-armed bandit for each gene of the genome that can take on $m$ possible values.
Each bandit works by recording how many times each arm has been pulled, i.e., the number of evaluations of each arm, and the difference in empirical reward between the previous fitness of the genome
and the fitness obtained as a result of the selected mutation.



Note that instead of pulling an arm to gain some reward as in the normal bandit terminology, each bandit stores a state and has $m$ arms where each arm $i \in \{1,\dots,m\}$ stores the statistics of a transition: $T_i \in Transitions(S)$ with $|Transitions(S)|=m$. $Transitions(S)$ denotes the set of transitions at state $S$.
Thus, for a genome of $n$ genes, $n$ multi-armed bandits are created, assuming they are independent. 

Each bandit can have a different number of arms, depending on the problem and transition sets. In a $n$-dimensional OneMax problem or Royal Road function $R_1$, $n$ $2$-armed bandits are required.

For the rest of this paper we assume that $m=2$ i.e. we are dealing with binary strings,
though the extension to larger alphabets should be straightforward.

For any position (gene) at a given state $S$, there is one single possible action \emph{flip} and two transitions, the next state will be
\[
S'= 
\begin{cases}
    1,& \text{if it's } 0 \text{ at state } S\\
    0,& \text{otherwise.}
\end{cases}
\]

\subsection{Urgency}
At each iteration of the Bandit-based RMHC algorithm, the bandit agents manage the selection of the gene with maximal $urgency$ to mutate:
\begin{equation}\label{eq:Urgent}
i^{*} = \underset{i\in\{1,2,\dots,n\}}{\argmax} urgency_i.
\end{equation}

The urgency of each bandit is derived from the standard UCB equation, except that we invert the normal use of the exploitation term, i.e. the first term in the RHS of Equation~\ref{eq:UrgentUCB}.
Intuitively, this says that if a particular state of a bandit is already good, then it's value should not be changed.
The exploration term is there to ensure that as the total number of iterations $N_i$ increases, so occasionally an apparently poorer option will be tried.

For any $2$-armed bandit $i\in\{1,2,\dots,n\}$, the $urgency_i$ is defined as
\begin{equation} \label{eq:UrgentUCB}
urgency_i = - \underset{j\in\{0,1\}}\max \bar{\Delta}_i(j) + \sqrt{\frac{\log(N_i + 1)}{2N_i(j)}} + \U(1e^{-6}),
\end{equation}
where $N_i$ is the number of times the $i^{th}$ bit is selected; $N_i(j)$ is the number of times the state $j$ is reached when the $i^{th}$ bit is selected; $\bar{\Delta}_i(j)$ is the empirical mean difference between the fitness values if the state $j$ is reached when the $i^{th}$ bit is selected, i.e., the changing of fitness value; $\U(1e^{-6})$ denotes a uniformly distributed value between $0$ and $1e^{-6}$ which is used to randomly break ties.

This means that for each position in the bit string (i.e. for each gene) we have a simple bandit model that requires only 3 additional parameters for book-keeping: one parameter to model the fitness change when flipping a bit from one to zero, another one for the opposite flip, and one to count
the number of times that a bandit has been selected ($N_i(j)$).

\def\nouse{
At each generation of (1+1)-ES, the bandits agent manage the selection of the gene to mutate (Algo. \ref{algo:bandit}).
\begin{algorithm}
\caption{\label{algo:bandit}Bandit-based mutation. $\bar{\Delta}_i(j)$ is the empirical mean difference between the fitness values if the $i^{th}$ bit changed from $1-j$ to $j$, thus the reward of transition from $1-j$ to $j$. $\U(\epsilon)$ denotes a uniformly distributed value $\in (0,\epsilon)$.}
\begin{algorithmic}
\Function{$\pi_{Bandit}$}{$\mathbf{s} \in \{0,1\}^n$}
\State{$\mathbf{s'} \gets \mathbf{s}$}
\State{$i^{*} \gets \underset{i\in\{1,2,\dots,n\}}{\argmax} \{- \underset{j\in\{0,1\}}\max \bar{\Delta}_i(j) + \sqrt{\frac{\log(N + 1)}{2N_i}} + \U(\epsilon)\}$
}
\State{$s'_i \gets 1-s'_i$}\\
\Return{$\mathbf{s'}$}
\EndFunction
\end{algorithmic}
\end{algorithm}
}

\subsection{Noise-free case}
Algorithm \ref{algo:noisefree} presents the bandit-based RMHC in the noise-free case.
To solve a noise-free problem, no resampling is necessary if the evaluation number and fitness value of the best-so-far genome can be saved. It is worth noting that, for problems in which computing the fitness value is difficult or requires high computational cost, saving the fitness of a solution is far less expensive than re-evaluating it again. For the further work, we are interested in applying our proposed approach to more difficult problems (such as game level generation and evaluation~\cite{khalifageneral}).
Our main interest is in improving the speed of convergence to approximately optimal states.
\begin{algorithm}[btph]
\caption{\label{algo:noisefree}Bandit-based RMHC in the noise-free case.}
\begin{algorithmic}[1]
\Require{$n \in \N^*$: genome length}
\Require{$m \in \N^*$: dimension of search space}
\State{Randomly initialise a genome $\mathbf{x} \in \R^m$}
\State{$bestFitSoFar \gets fitness(\mathbf{x})$}
\State{$N \gets 1$}	\Comment{Total evaluation number}
\While{time not elapsed}
    \State{Select the element $i^*$ to mutate using Eqs. \ref{eq:Urgent} and \ref{eq:UrgentUCB}}
    \State{$\mathbf{y} \gets$ after mutating the element $i^*$ of $\mathbf{x}$}
    \State{$Fit_{\mathbf{y}} \gets fitness(\mathbf{y})$}
    \State{$N \gets N + 1$}\Comment{Update the counter}
    \If{$Fit_{\mathbf{y}} \geq bestFitSoFar$}
    	\State{$\mathbf{x} \gets \mathbf{y}$}	\Comment{Update the best-so-far genome}
        \State{$bestFitSoFar \gets Fit_{\mathbf{y}}$}
    \EndIf
\EndWhile
\State{\Return{$\mathbf{x}$}}
\end{algorithmic}
\end{algorithm}

\subsection{Noisy case}\label{sec:noisy}
We now consider the noisy case.
The Bandit-based RMHC in the noisy case is formalised in Algorithm \ref{algo:noisy}.
In the noisy case, the best-so-far genome requires multiple evaluations to reduce the effect of noise, this is called resampling.

The statistics of the best-so-far genome are stored, thanks to which, instead of comparing directly the fitness values of the offspring to the one of the best-so-far genome, the average fitness value of the best-so-far genome is compared at each generation. Therefore, the computational cost involved in the evaluation of the genome determines the computational cost of this algorithm.

Resampling has been proved to be a powerful tool to improve the local performance of EAs in noisy optimization~\cite{arnold2006general,beyer2013theory} and a variety of resampling rules applied to EAs in continuous noisy optimization are studied in \cite{liu2015portfolio}.
Interestingly, Qian et al.~\cite{qian2014effectiveness} proved theoretically and empirically that under some conditions, resampling is not beneficial for a (1+1)-EA optimizing $10$-OneMax under additive Gaussian noise. 
This is contrary to our findings, though we use much larger bit string lengths for our
main results (from $50$ to $1000$, unlike the length of $10$ used by Qian et al.) and different algorithm (RMHC using $1$-bit mutation, unlike (1+1)-EA where every bit could mutate with a probability), 
and we also found resampling to not improve results for small strings (e.g. for $N=10$).

\begin{algorithm}[btph]
\caption{\label{algo:noisy}Bandit-based RMHC in the noisy case.}
\begin{algorithmic}[1]
\Require{$n \in \N^*$: genome length}
\Require{$m \in \N^*$: dimension of search space}
\State{Randomly initialise a genome $\mathbf{x} \in \R^m$}
\State{$bestFitSoFar \gets fitness(\mathbf{x})$}
\State{$M \gets 1$}	\Comment{Evaluation number of the best-so-far genome}
\State{$N \gets 1$}	\Comment{Total evaluation number}
\While{time not elapsed}
    \State{Select the element $i^*$ to mutate using Eqs. \ref{eq:Urgent} and \ref{eq:UrgentUCB}}
    \State{$\mathbf{y} \gets$ after mutating the element $i^*$ of $\mathbf{x}$}     \State{$Fit_{\mathbf{x}} \gets fitness(\mathbf{x})$}
    \State{$Fit_{\mathbf{y}} \gets fitness(\mathbf{y})$}
    \State{$N \gets N + 2$}	\Comment{Update the counter}
    \State{$averageFitness \gets \frac{bestFitSoFar*M+Fit_{\mathbf{x}}}{M+1}$}
    \If{$Fit_{\mathbf{y}} \geq averageFitness$}
    	\State{$\mathbf{x} \gets \mathbf{y}$}	\Comment{Update the best-so-far genome}
   		\State{$bestFitSoFar \gets Fit_{\mathbf{y}}$}
        \State{$M \gets 1$}
    \Else
    	\State{$bestFitSoFar \gets averageFitness$}
        \State{$M \gets M + 1$}
    \EndIf
\EndWhile
\State{\Return{$\mathbf{x}$}}
\end{algorithmic}
\end{algorithm}

\def\later{
Let $p_{TA}$, $p_{FA}$, $p_{TR}$ and $p_{FR}$ denote the probability of true acceptance, false acceptance, true rejection and false rejection, respectively (Fig. \ref{fig:state}).
\begin{figure}[h]
\centering
\includegraphics[width=.8\linewidth]{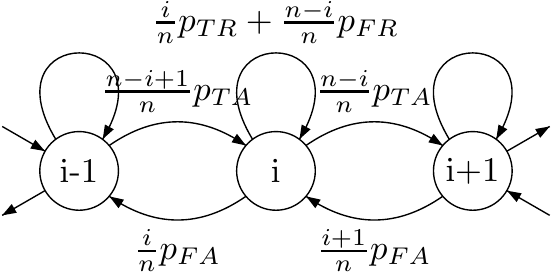}
\caption{\label{fig:state}Game states and transitions in a noisy OneMax problem.}
\end{figure}

Consider a $n$-bit OneMax problem with constant variance noise.
At state $i$, if the newly generated genome $y$ is a better than the current genome $x$, i.e. $f(y) > f(x)$, when comparing two genomes without resampling or storing the statistics of the best-so-far genome, the probability of true acceptance is
\begin{eqnarray*}
p_{TA} &=& \P(f'(y) > f'(x) | f(y)=i+1, f(x)=i)\\
&=& \P(\omega_y + 1 > \omega_x)\\
&=& \P(\omega_y - \omega_x > -1).
\end{eqnarray*}
As $\omega_y$ and $\omega_x$ and independent samples of $\G(0,1)$, $\omega_y - \omega_x$ $\sim \G(0,\sqrt2)$. Then,
\begin{eqnarray*}
p_{TA} &=& \P(\omega_y - \omega_x > -1) \\
&=& \P(\omega_y - \omega_x \leq 1) \\
&=& CDF_{Gaussian(0,\sqrt2)}(1) \\
&=& \frac12 + \frac12 erf(\frac12).
\end{eqnarray*}

When comparing two genomes using the statistics of the best-so-far genome:
\begin{equation}
p_{TA} = \frac12 + \frac12 erf(\sqrt{\frac{M+1}{2M+4}}) \geq \frac12 + \frac12 erf(\frac12).
\end{equation}
Equality holds when $M=1$, $M$ counts how many times the best-so-far genome is evaluated.

When comparing two genomes using $r>1$ resamplings and the statistics of the best-so-far genome:
\begin{equation}
p_{TA} = \frac12+\frac12erf(\sqrt{\frac{r(M+1)}{2M+4}}).
\end{equation}

Therefore, the probability of true acceptance is improved by resampling genomes and storing the statistics of the best-so-far genome. 

To simplify, we only consider the resampling in the theoretical analysis (no statistic is stored, $M=0$), thus, 
\begin{equation}\label{eq:ta}
p_{TA} = \frac12+\frac12erf(\frac{\sqrt{r}}{2}).
\end{equation}
Respectively,
\begin{eqnarray}
p_{TR} &=& \frac12+\frac12erf(\frac{\sqrt{r}}{2}) = p_{TA},\label{eq:tr}\\
p_{FA} &=& 1-p_{TA},\label{eq:fa}\\
p_{FR} &=& 1-p_{TR}.\label{eq:fr}
\end{eqnarray}

When using a RMHC which uniformly randomly selects the bit to mutate, the probability of transiting from state $i$ to $i+1$ is $\frac{n-i}{n}p_{TA}$ shown in Fig. \ref{fig:state}.
So the total expected number of fitness evaluations needed to get from the state $0$ to the state $n$ is computed as follows:
\begin{eqnarray}
N &=& \sum_{i=0}^{n-1} \frac{n}{(n-i)p_{TA}}\\
&=& \frac{1}{p_{TA}}\sum_{i=0}^{n-1} \frac{n}{n-i}.
\end{eqnarray}
For a noise-free $n$-bit OneMax problem, the expected number of fitness evaluations is $\sum_{i=0}^{n-1} \frac{n}{n-i} \in O(n\log{n})$.
For a $n$-bit OneMax problem with constant variance noise, the expected number of fitness evaluations $\frac{1}{p_{TA}}\sum_{i=0}^{n-1} \frac{n}{n-i}$ is of same order but with a higher constant $C$ which is inversely proportional to $p_{TA}$. 
Bigger $r$ leads to a higher $p_{TA}$, consequently a lower upper bound for the expected number of fitness evaluations.
}

\def\tocheck{
\paragraph{Version 2}
When using a RMHC which uniformly randomly selects the bit to mutate, using the transition probabilities shown in Fig. \ref{fig:state}, we obtain the following expected number of evaluations needed to transit from state $i$ to state $i+1$ is :
\begin{eqnarray*}
\E N(i,i+1) &=& 
\frac{n-i}{n}p_{TA}\\
&&+ (1+\E N(i,i+1))(\frac{i}{n}p_{TR} + \frac{i-1}{n}p_{FR}) \\
&&+ (2+\E N(i,i+1))\frac{i}{n}p_{FA}\frac{n-i+1}{n}p_{TA}.
\end{eqnarray*}

To simplify the notation, from now on, we use $\tilde N_i$ to denote $\E N(i,i+1)$, the expected number of evaluations needed to transit from state $i$ to state $i+1$; and $p=\frac12+\frac12erf(\frac{\sqrt{r}}2)$. 
By Eqs. \ref{eq:ta}, \ref{eq:tr}, \ref{eq:fa} and \ref{eq:fr},
\begin{eqnarray*}
N_i &=& 
\frac{n-i}{n}p\\
&&+ (1+N_i)(\frac{i}{n}p + \frac{i-1}{n}(1-p)) \\
&&+ (2+N_i)\frac{i}{n}(1-p)\frac{n-i+1}{n}p.
\end{eqnarray*}
\begin{equation}
\E N(i,i+1) = A \E N(i-1,i) + B,
\end{equation}
with $A=\frac{i(1-p_{TA})}{i+(n-2i)p_{TA}}$ and $B=-\frac{i}{i+(n-2i)p_{TA}}$.
}

\section{Experimental results}
We apply first our proposed algorithm on the OneMax problem and the Royal Road function $R_1$ in a noise-free case, and then evaluate the performance of our algorithm on the OneMax problem with the presence of noise.
Each experiment is repeated $100$ times using randomly initialised strings.

\subsection{OneMax}
\begin{figure}
\centering
\includegraphics[width=\linewidth]{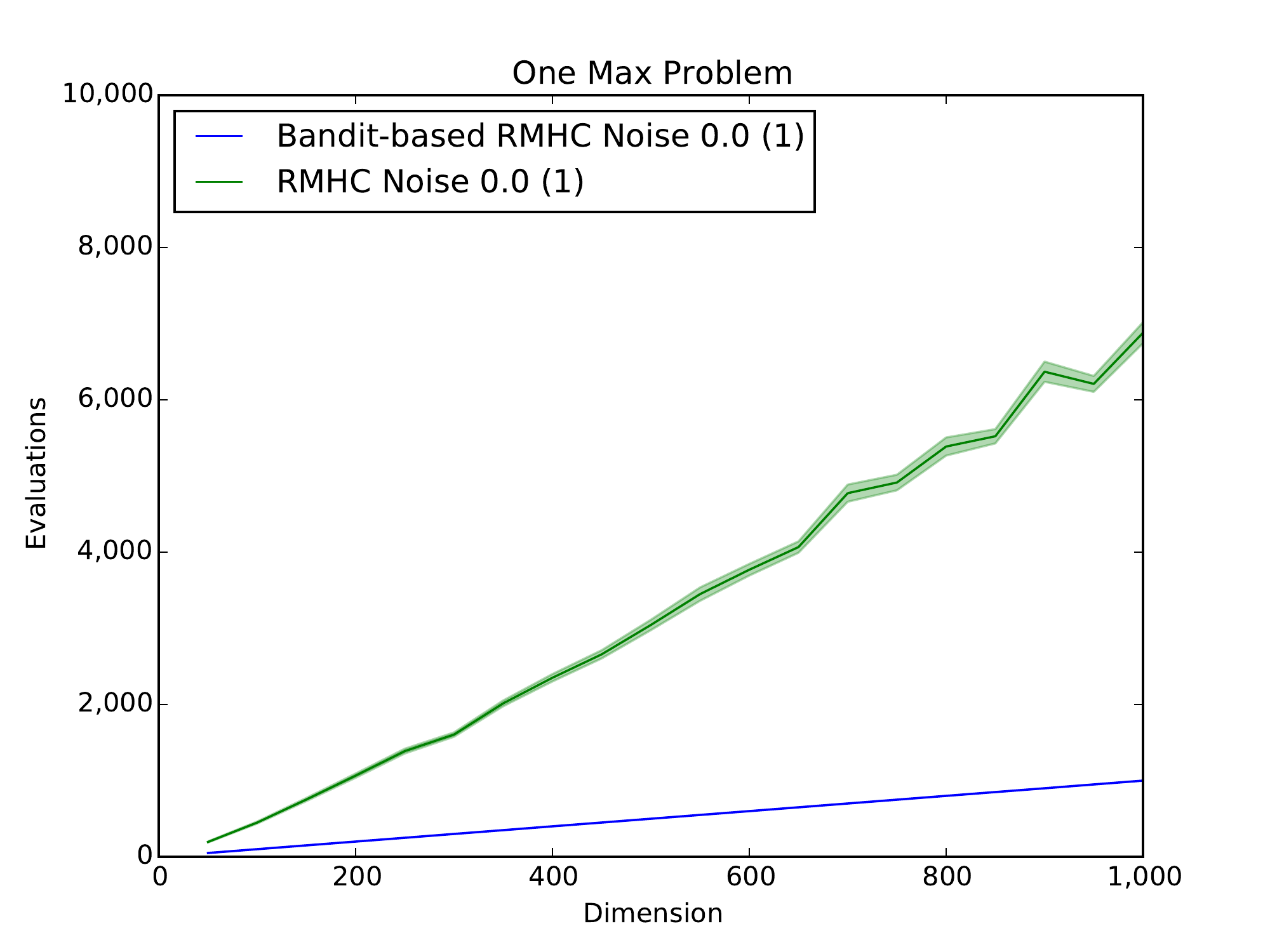}
\caption{\label{fig:one}Average evaluations required to find the optimal solution in the noise-free OneMax problem.}
\end{figure}

The results in noise-free OneMax problem of different dimensions is presented in Fig. \ref{fig:one}. 
In the noise-free case, the average fitness evaluations used by bandit-based RMHC to solve the problem is close to the problem dimension, while the original RMHC required approximate $5$ times more budget.
\def\nouse{
\begin{table}
\centering
\caption{\label{tab:one}Average evaluations required to find the optimal solution in the noise-free OneMax problem.}
\begin{tabular}{ccc}
\hline
\multirow{2}{*}{Dimension} & \multicolumn{2}{c}{Average evaluations $\pm$ standard error}\\
\cline{2-3}
 & RMHC & Bandit-based RMHC\\
\hline
50 & 194.49 $\pm$ 6.69 & 49.95 $\pm$ 0.10\\                  
100 & 449.47 $\pm$ 13.53 & 100.02 $\pm$ 0.13\\
150 & 752.23 $\pm$ 20.75 & 150.18 $\pm$ 0.13\\
200 & 1065.38 $\pm$ 27.56 & 200.03 $\pm$ 0.13\\
250 & 1387.73 $\pm$ 31.11 & 249.70 $\pm$ 0.18\\
300 & 1604.79 $\pm$ 28.37 & 300.14 $\pm$ 0.12\\
350 & 2015.00 $\pm$ 37.62 & 349.82 $\pm$ 0.15\\
400 & 2350.96 $\pm$ 48.50 & 400.19 $\pm$ 0.14\\
450 & 2658.32 $\pm$ 54.15 & 449.82 $\pm$ 0.17\\
500 & 3044.25 $\pm$ 68.43 & 499.74 $\pm$ 0.16\\
550 & 3449.32 $\pm$ 87.86 & 549.79 $\pm$ 0.17\\
600 & 3769.22 $\pm$ 76.49 & 599.77 $\pm$ 0.17\\
650 & 4066.74 $\pm$ 74.90 & 649.87 $\pm$ 0.16\\
700 & 4774.75 $\pm$ 112.98 & 699.99 $\pm$ 0.18\\
750 & 4914.25 $\pm$ 102.39 & 749.96 $\pm$ 0.13\\
800 & 5386.99 $\pm$ 119.23 & 800.18 $\pm$ 0.10\\
850 & 5521.65 $\pm$ 93.89 & 850.05 $\pm$ 0.13\\
900 & 6369.32 $\pm$ 132.50 & 899.65 $\pm$ 0.14\\
950 & 6209.33 $\pm$ 104.56 & 950.16 $\pm$ 0.11\\
1000 & 6880.20 $\pm$ 138.17 & 1000.07 $\pm$ 0.15\\
\hline
\end{tabular}
\end{table}
}

\begin{figure*}[btph]
\centering
\begin{subfigure}[b]{0.45\textwidth}
\includegraphics[width=\textwidth]{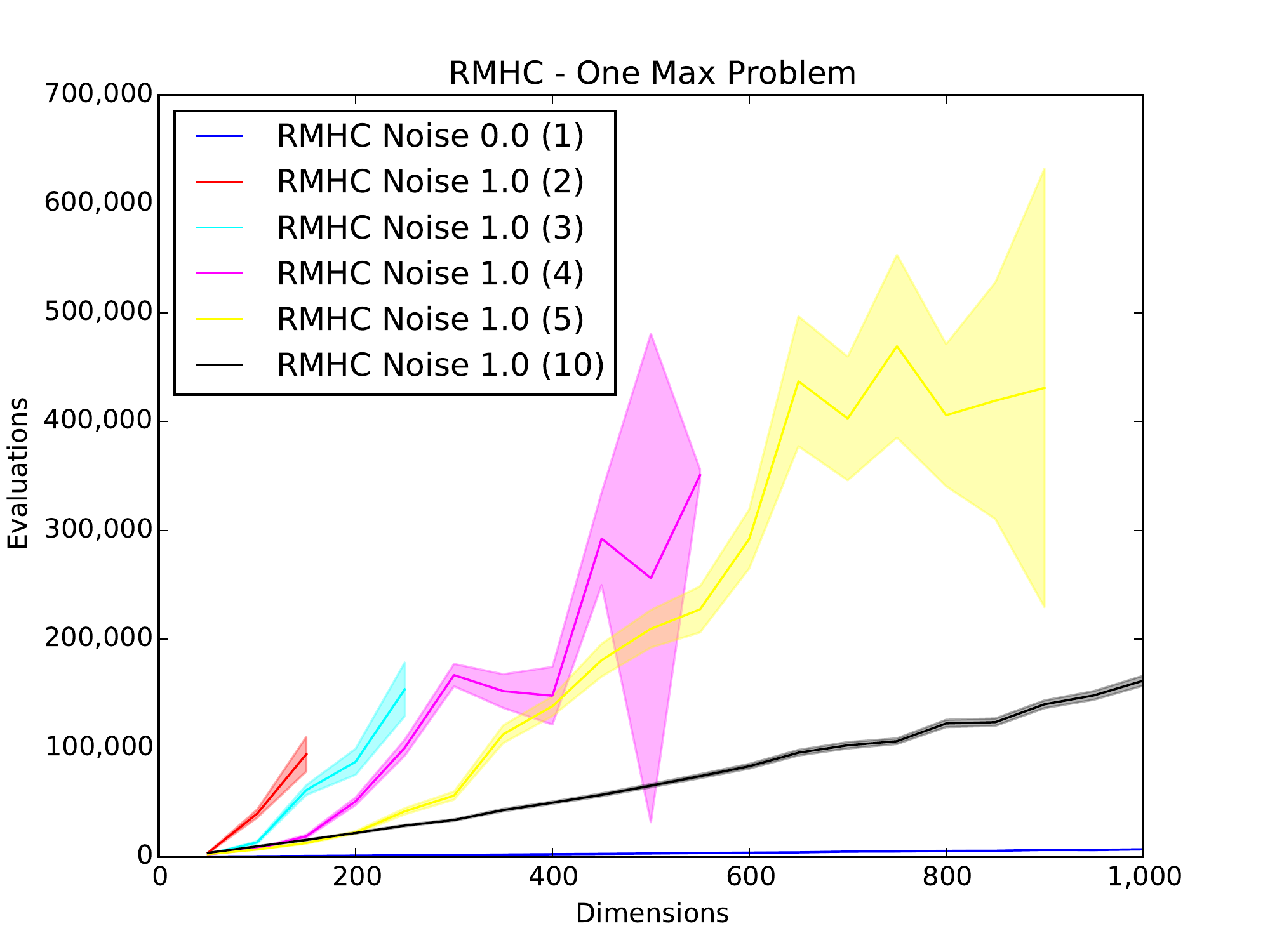}
\caption{\label{fig:rmhc1}Average evaluations used by RMHC.}
\end{subfigure}
\begin{subfigure}[b]{0.45\textwidth}
\includegraphics[width=\textwidth]{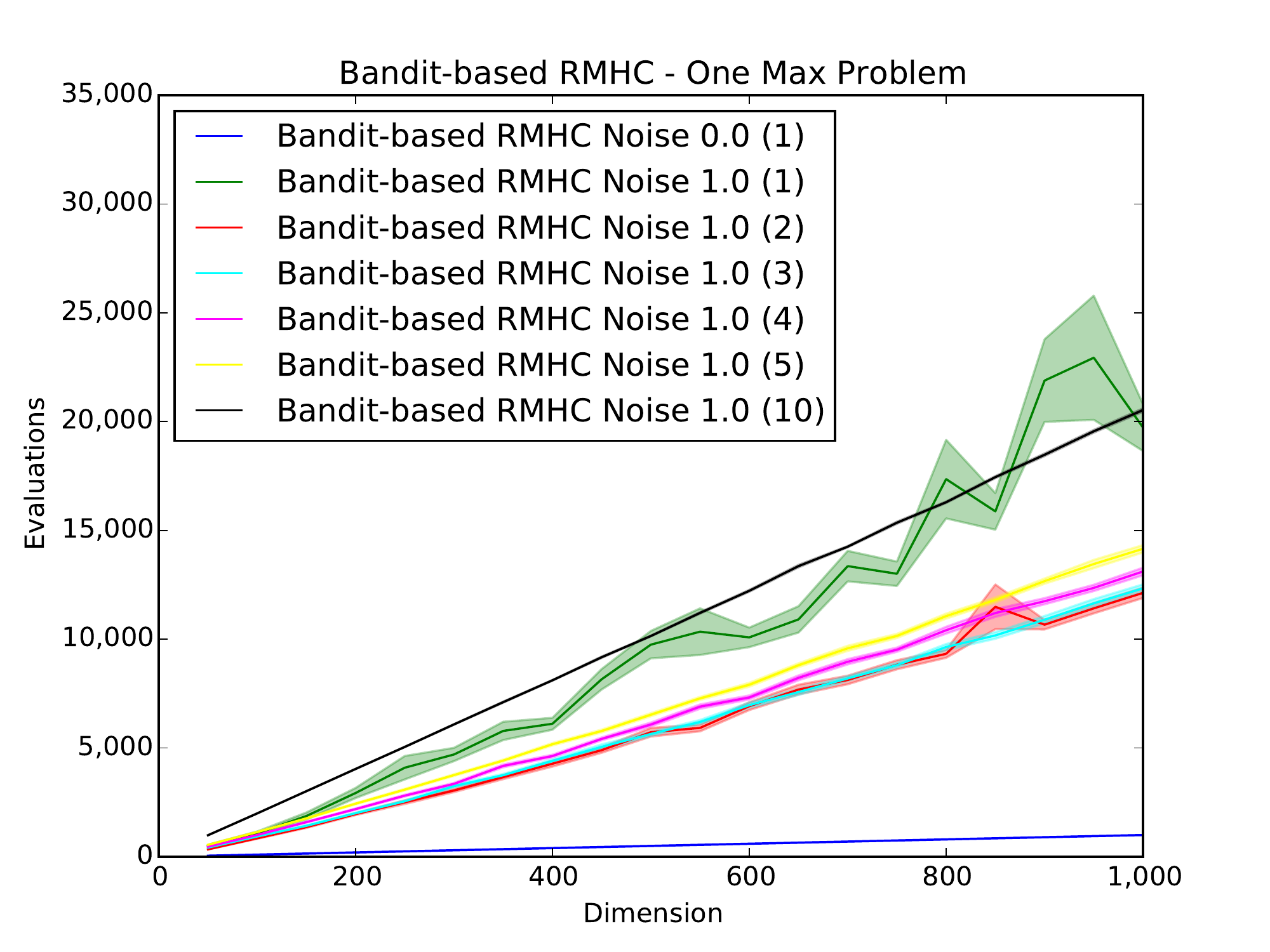}
\caption{\label{fig:banditea1}Average evaluations used by bandit-based RMHC.}
\end{subfigure}
\begin{subfigure}[b]{0.45\textwidth}
\includegraphics[width=\textwidth]{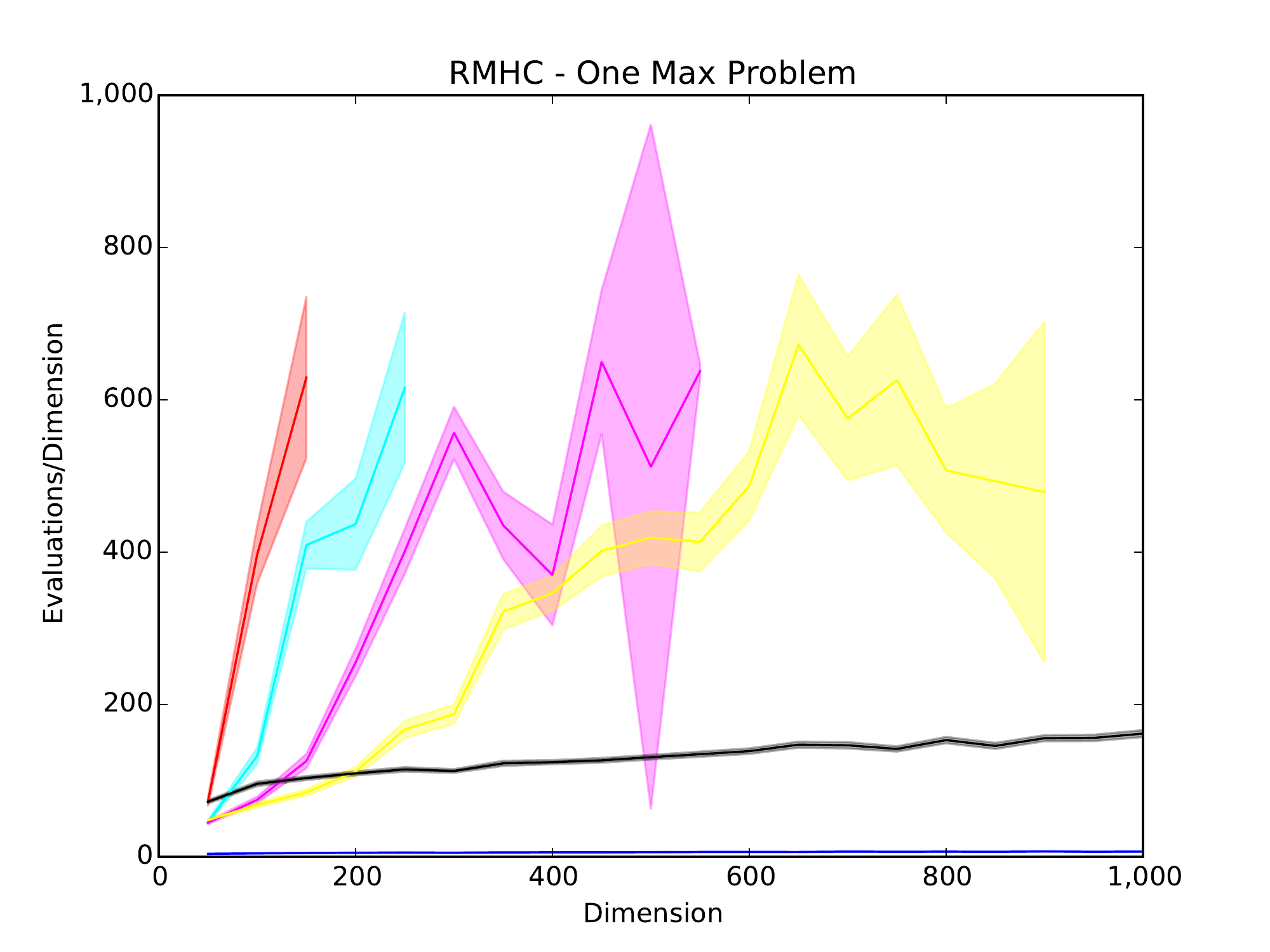}
\caption{\label{fig:rmhc2}RMHC. \emph{y-axis: average \#evals$/$dim}.}
 \end{subfigure}
\begin{subfigure}[b]{0.45\textwidth}
\includegraphics[width=\textwidth]{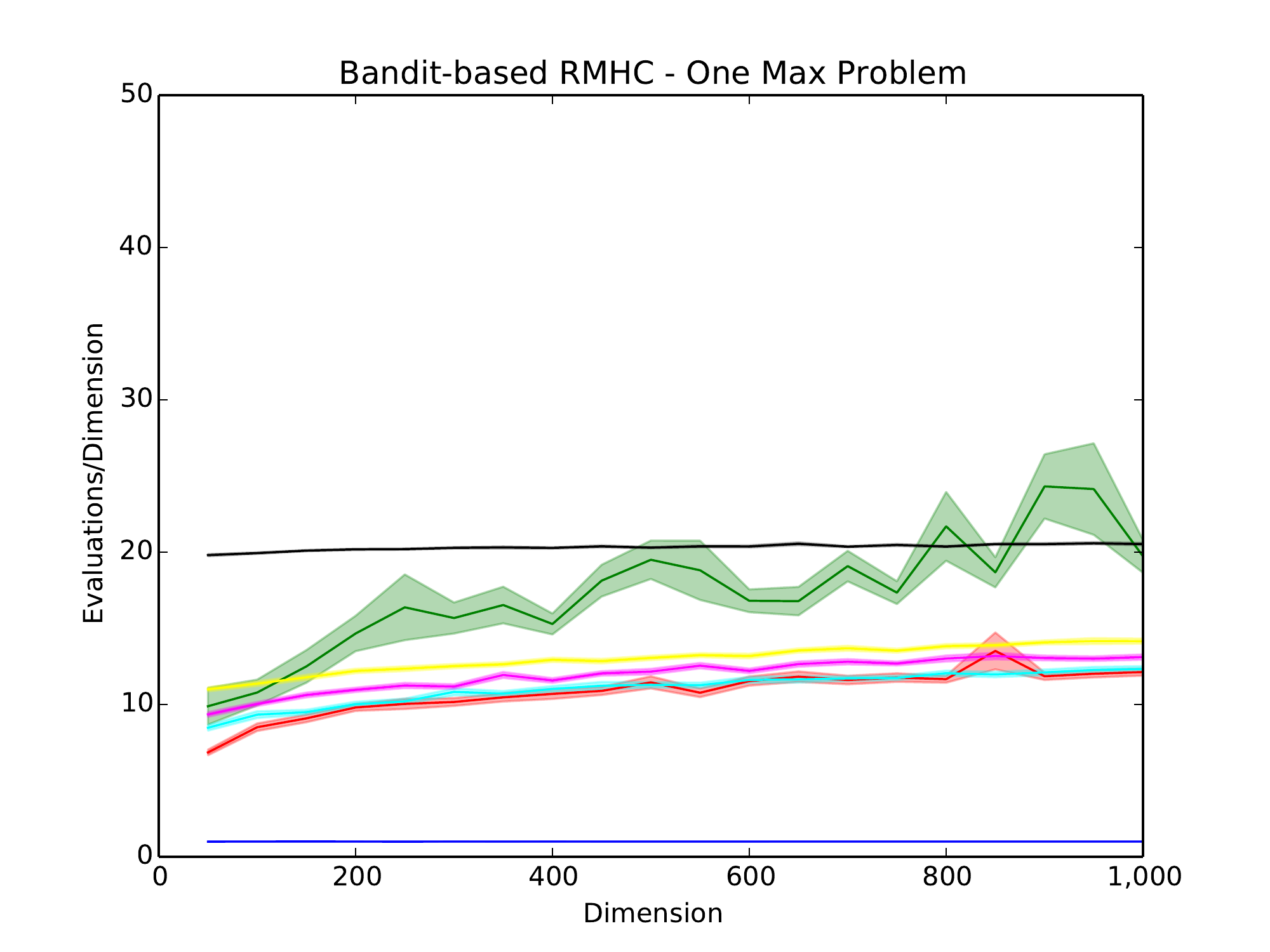}
\caption{\label{fig:banditea2}Bandit-based RMHC. \emph{y-axis: average \#evals$/$dim.}}
\end{subfigure}
\caption{\label{fig:allGames}Performance of RMHC (left) and bandit-based RMHC (right) in the OneMax problem, without noise and with constant variance noise ($\sim \G($0$,$1$)$). The y-axis range of RMHC ((a) and (c)) is {\textbf{20 times}} the ones of bandit-based RMHC ((b) and (d)). The resampling number in the noisy case is given between brackets. The standard error is shown as well as a faded area around the average. In the noisy case, our proposed bandit-based RMHC significantly outperforms RMHC. In the noisy case, the RMHC without resampling is not shown in because it could not solve the problem within  $(1000 \times Dimension)$ function evaluations.}
\end{figure*}

Fig. \ref{fig:allGames} illustrates the empirical average number of fitness evaluations required to reach the optimum value using RMHC and bandit-based RMHC in the OneMax problem of different dimensions with constant variance noise ($\sim \G($0$,$1$)$). The resampling number in the noisy case is given between brackets. For comparison, the results in noise-free OneMax is also included (blue curves).

As is exhibited in the graph, with the presence of constant variance noise:
\begin{itemize}
\item Using RMHC, larger resampling number ($10$) leads to a faster convergence to the optimum (Fig. \ref{fig:rmhc1}) on high-dimension problems; when the problem dimension is low, resampling number equals to $3$, $4$ or $5$ leads to a faster convergence to the optimum (Fig. \ref{fig:rmhc2}); the ratio of average fitness evaluations to the problem dimension increases noticeably when a small resampling number is used (Fig. \ref{fig:rmhc2}).
\item Using bandit-based RMHC, the ratio of average fitness evaluations to the problem dimension remains stable when resampling is used (Fig. \ref{fig:banditea2}); the total evaluation number scales almost linearly with the problem dimension; the optimal resampling number is $2$ (red curve).
\item As can be seen clearly from the figures, for an identical OneMax problem with the presence of noise, our proposed bandit-based RMHC significantly outperforms the standard RMHC by a 
very large margin - in some cases requiring a factor of ten fewer fitness evaluations.
\end{itemize}

\subsection{Royal Road}
Fig. \ref{fig:royal} shows the empirical average fitness evaluations required to find the optimum of the noise-free Royal Road function using RMHC and bandit-based RMHC, respectively. 
\begin{figure*}[btph]
\centering
\begin{subfigure}[b]{0.45\textwidth}
\includegraphics[width=\textwidth]{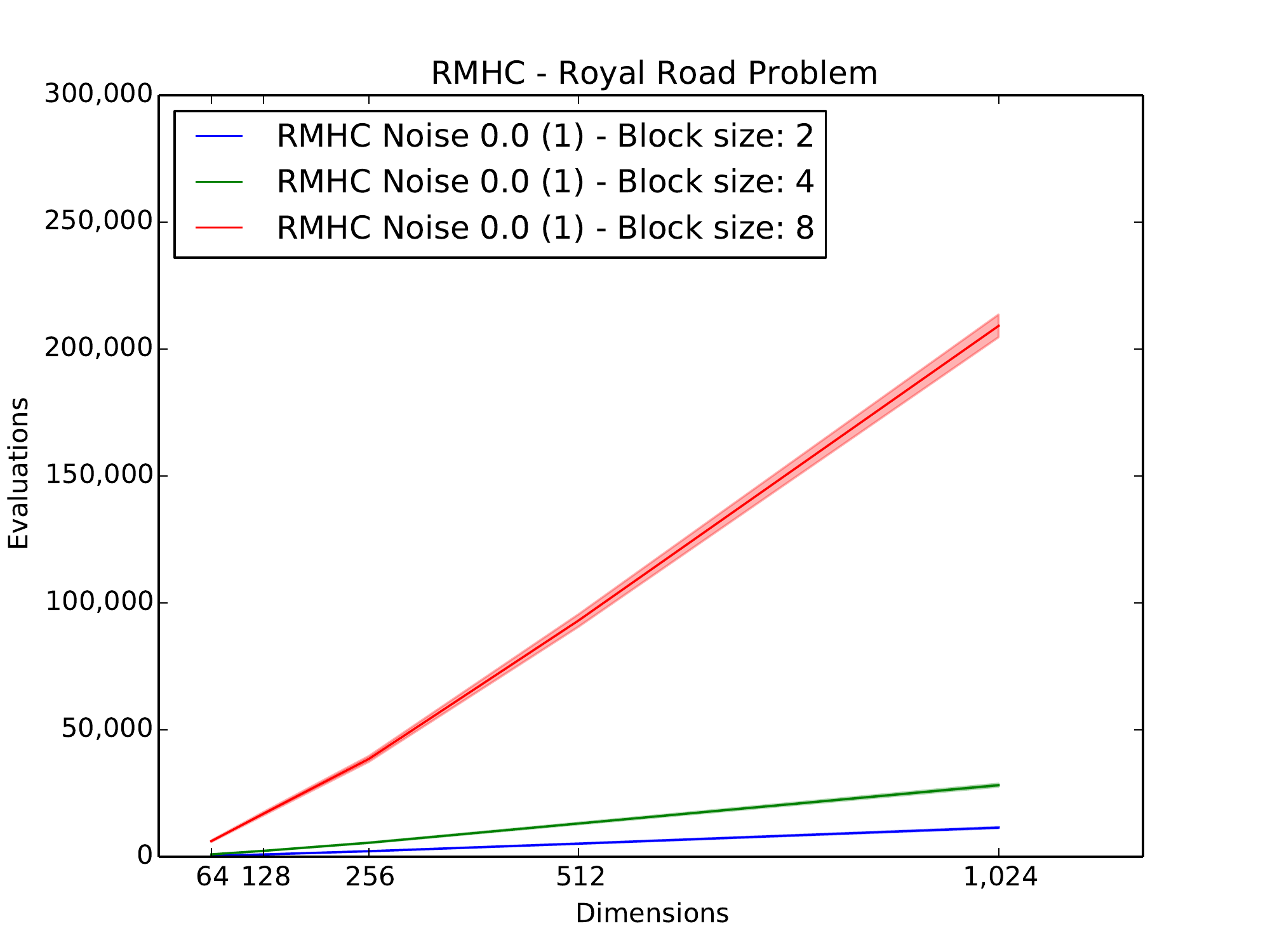}
\caption{\label{fig:rrrmhc1}Average evaluations used by RMHC.}
 \end{subfigure}
\begin{subfigure}[b]{0.45\textwidth}
\includegraphics[width=\textwidth]{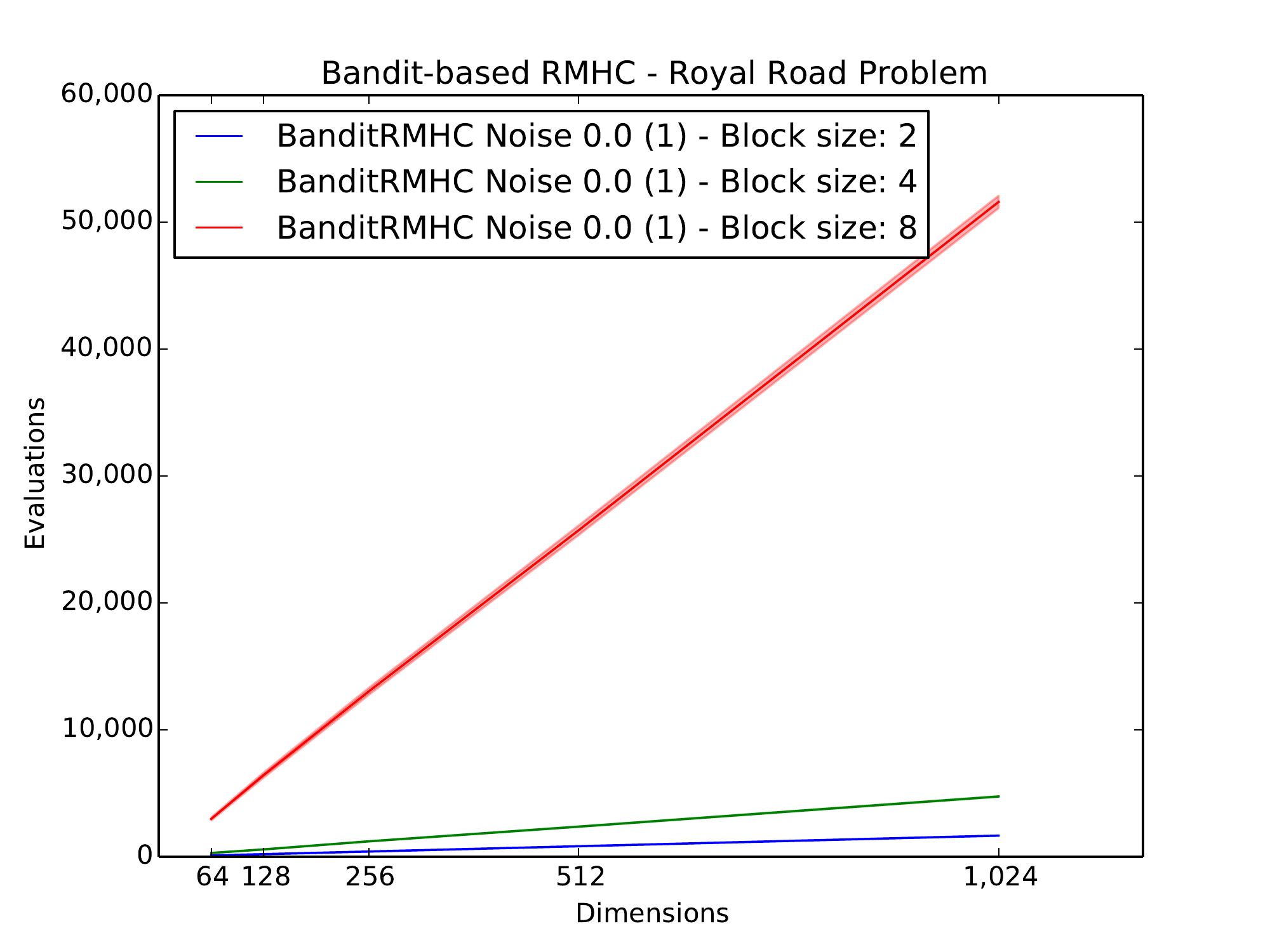}
\caption{\label{fig:rrbanditea1}Average evaluations used by bandit-based RMHC.}
\end{subfigure}
\begin{subfigure}[b]{0.45\textwidth}
\includegraphics[width=\textwidth]{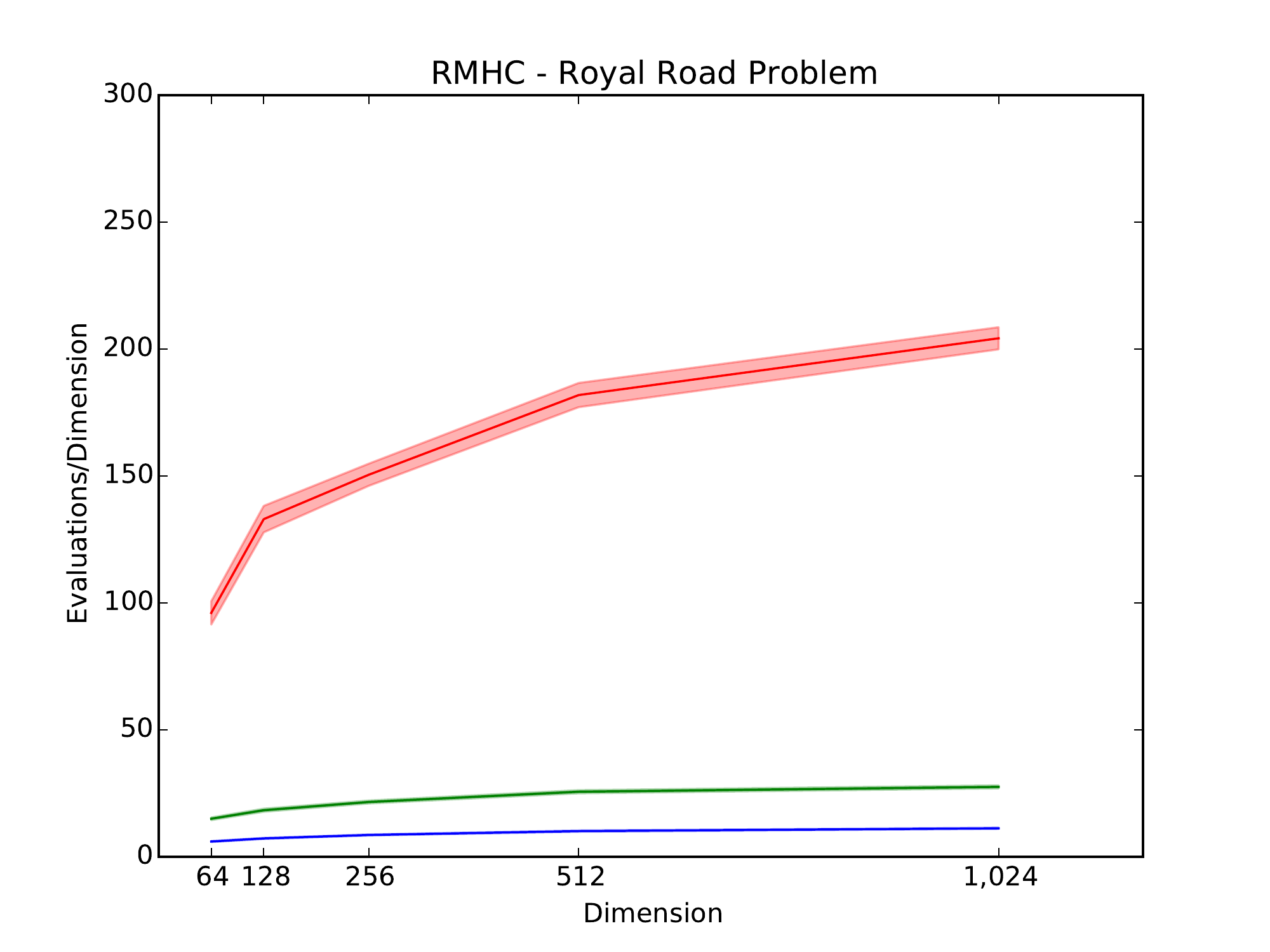}
\caption{\label{fig:rrrmhc2}RMHC. \emph{y-axis: average \#evals$/$dim}.}
 \end{subfigure}
\begin{subfigure}[b]{0.45\textwidth}
\includegraphics[width=\textwidth]{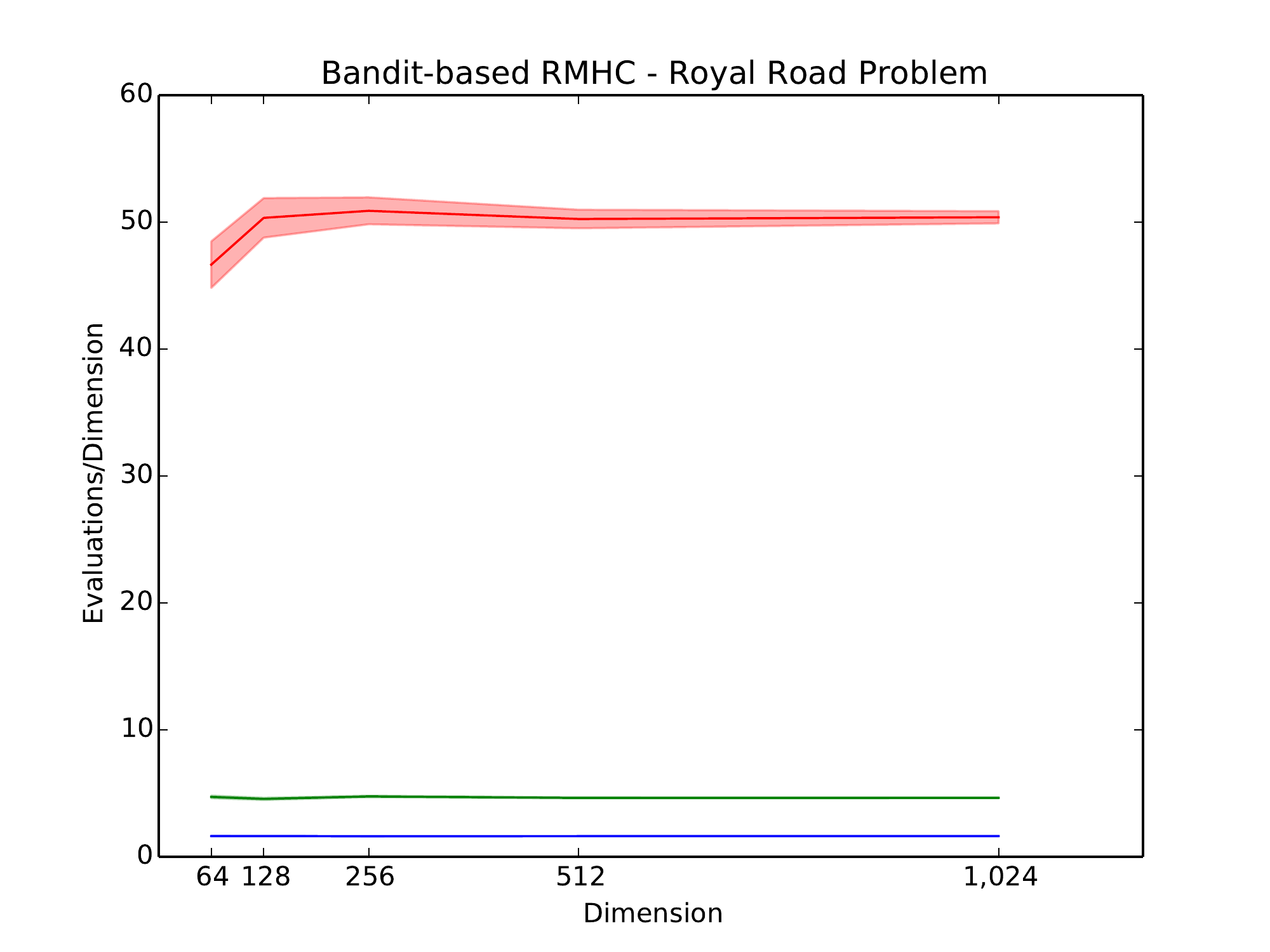}
\caption{\label{fig:rrbanditea2}Bandit-based RMHC. \emph{y-axis: average \#evals$/$dim}.}
\end{subfigure}
\caption{\label{fig:royal}Performance of RMHC (left) and bandit-based RMHC (right) in the noise-free Royal Road problem with different block sizes. Note that the y-axis range of RMHC in (a) and (c) is {\textbf{5 times}} the ones of bandit-based RMHC in (b) and (d). The standard error is also given as a faded area around the average.}
\end{figure*}

For a fixed length, bigger block size results in a harder problem and more fitness evaluations.
The ratio of average fitness evaluations to the problem dimension increases with the problem dimension when a small resampling number is used (Fig. \ref{fig:rmhc2}).
For an identical problem, bandit-based RMHC required far fewer fitness evaluations than RMHC to find the optimal solution. It can be seen from the curves that for an identical block size, using bandit-based RMHC, the total evaluation number scales linearly with the problem dimension.

In addition, to find the optimum string of $8$ blocks of size $8$, our  bandit-based RMHC used half number of function evaluations than the RMHC used by Mitchell et al. \cite{mitchell1993will}, which was the most efficient algorithm in their experiments.


The Royal Road functions involve a harder credit assignment~\cite{sutton1984temporal} problem than standard OneMax, an important aspect of sequential decision making. The reward for correctly mutating a bit 
is usually delayed, and dependent on many other correct bit settings.

Regarding credit assignment within the algorithm, the bandit-based RMHC uses \emph{urgency} (Eq. \ref{eq:UrgentUCB})  to model this, by attempting to track the fitness gained when
switching a gene to a particular value.  More use is made of the available information, leading to faster learning (see \cite{lucas2008investigating} and \cite{lucas2010estimating} for more analysis of the information rates of simple evolutionary algorithms).

If this information was exploited in a way that was too na\"\i ve or too greedy, this could lead the algorithm to rapidly become stuck on poor values, especially for the noisy problems tested in this paper. However, the exploration term naturally counteracts such tendencies.

\section{Conclusion}

This paper presented the first bandit-based Random Mutation Hill Climber (RMHC) - a simple but effective type of evolutionary algorithm. The algorithm was compared with the standard RMHC on the OneMax problem and Royal Road function. Tests were also made using a noisy OneMax problem together with resampling in each algorithm to ameliorate the effects of the noise. 

On noise-free and noisy OneMax problems and Royal Road function, our bandit-based RMHC algorithm significantly outperforms the RMHC, in some cases using a factor of ten fewer evaluations in the noisy case.

Furthermore, the fitness evaluations required by the bandit-based RMHC finding the optimal solution is an approximately linear function of the problem dimension when the resampling number is $2$.
For an identical Royal Road function $8$ blocks of size $8$), the bandit-based RMHC used half the number of function evaluations than the RMHC used by Mitchell et al. in \cite{mitchell1993will}, which was the most efficient algorithm in their experiments.

We obtain very promising results using this simple but effective bandit-based RMHC. The algorithm is designed for a large set of discrete optimisation problems where each fitness evaluation is expensive and the fitness is possibly noisy due to some uncertainties, which is quite common in real-world applications.

\section*{Further work}
The main work in progress is the theoretical analysis on expected evaluation number and accuracy, and the tests on noisy Royal Road problems.
Also, because the method makes such efficient use of 
the fitness evaluations, it is ready to be applied to expensive optimisation problems such as game level design evaluation~\cite{khalifageneral}.



%
\balance
\bibliographystyle{IEEEtran}
\bibliography{banditRMHC}
\end{document}